%% file: main_arxiv.tex
\documentclass[11pt]{article}
\usepackage[preprint]{acl}
\usepackage{times}
\usepackage{latexsym}
\usepackage[T1]{fontenc}
\usepackage[utf8]{inputenc}
\usepackage{microtype}
\usepackage{inconsolata}
\usepackage{amsmath,amssymb}
\usepackage{algorithm}
\usepackage{algorithmic}
\usepackage{graphicx}
\usepackage{booktabs}
\usepackage{multirow}
\usepackage{array}
\usepackage{tabularx}
\usepackage{placeins}
\usepackage{colortbl}
\definecolor{heatblue}{RGB}{35,98,156}
\newcommand{\heatcell}[2]{\cellcolor{heatblue!\the\numexpr24+76*(#1-14)/78\relax!white}\ifnum#1>65\textcolor{white}{#2}\else #2\fi}
\newcommand{\backbonelogo}[2]{\raisebox{-0.22\height}{\includegraphics[height=#1]{#2}}}
\usepackage{tcolorbox}
\newtcolorbox{promptbox}[2][]{
  colback=white,colframe=black!40,colbacktitle=black!55,coltitle=white,
  title={#2},fonttitle=\small\bfseries,fontupper=\small\ttfamily,
  boxrule=0.4pt,arc=1.5pt,left=7pt,right=7pt,top=6pt,bottom=6pt,
  before skip=8pt,after skip=8pt,before upper={\raggedright\setlength{\parskip}{4pt}},
  toptitle=3pt,bottomtitle=3pt,#1
}
\definecolor{cardigreen}{RGB}{233,247,234}
\usepackage{balance}
\usepackage{stfloats}
\newcommand{\Description}[1]{}

\AtBeginDocument{%
  }

\begin{document}
\raggedbottom

\title{Beyond Uncertainty Reduction:\\Calibrated Planning for Interactive Diagnosis}

\author{
\normalfont
Yue Zhou\textsuperscript{1,*}, Haiyang Zhou\textsuperscript{1,*},
Jin Zhang\textsuperscript{1}, Kong Wang\textsuperscript{2},\\
Zihao Rao\textsuperscript{3}, Maolin Wang\textsuperscript{4},
Hanwen Du\textsuperscript{5,\textdagger}, Youhua Li\textsuperscript{4,\textdagger}\\[4pt]
{\small\textsuperscript{1}Lanzhou University, China\quad
\textsuperscript{2}Hunan University, China}\\
{\small\textsuperscript{3}Chongqing University of Posts and Telecommunications, China}\\
{\small\textsuperscript{4}City University of Hong Kong, Hong Kong, China}\\
{\small\textsuperscript{5}The Ohio State University, USA}\\[4pt]
{\small\textsuperscript{*}Equal contribution.\quad
\textsuperscript{\textdagger}Corresponding authors.}\\
{\small\texttt{du.1128@osu.edu}, \texttt{youhuali2-c@my.cityu.edu.hk}}
}

\maketitle

\begin{abstract}
Interactive medical diagnosis requires an agent to gather patient evidence through sequential questions before committing to a diagnosis. Large language models (LLMs) now make such consultations feasible, and existing approaches typically select questions by how much they reduce uncertainty among candidate diseases. However, this objective is insensitive to the unequal clinical costs of diagnostic errors and omitted severe alternatives, so minimizing uncertainty need not minimize final diagnostic loss. To bridge this gap, we propose \textbf{CaRDI} (\emph{Calibrated Risk for Diagnostic Interaction}) , a planning framework built around a clinical loss combining primary diagnosis error with severity-weighted differential omission. Diagnostic and safety beliefs estimate the two loss components, while conformal residual calibration corrects for underestimation from partial evidence. The calibrated risk then guides question selection and stopping by balancing expected risk reduction against interaction cost. Experiments on multiple medical benchmarks show that CaRDI improves accuracy while requiring fewer questions. Further analysis shows that replacing entropy with clinical risk as the planning objective reduces clinical loss by 21.5\% and decreases severe omissions.  Our code is available at \textcolor{blue}{\url{https://anonymous.4open.science/r/CaRDI-A940/}}.
\end{abstract}

\input{sections/intro_empirical}

\begin{figure*}[t]
\centering
\includegraphics[width=\textwidth]{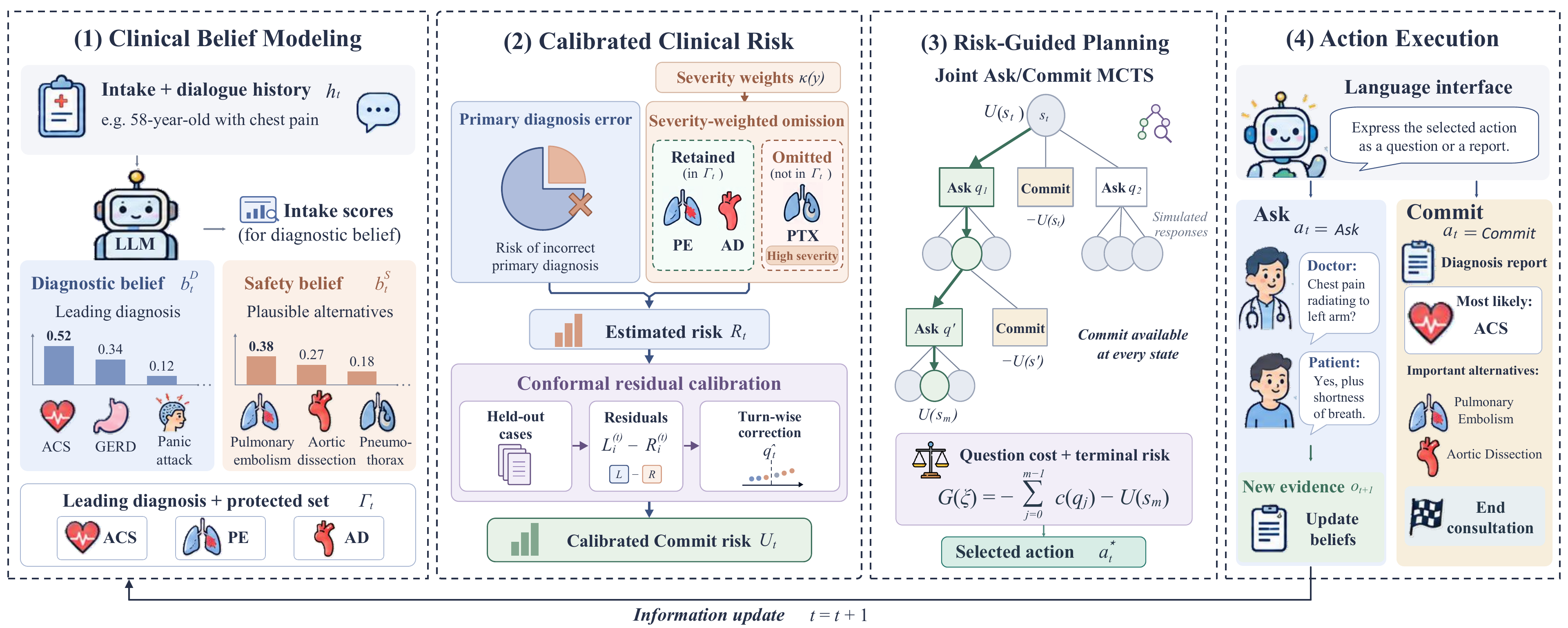}
\caption{Overview of the CaRDI framework. (1) Diagnostic and safety beliefs identify the leading diagnosis and plausible alternatives. (2) Severity-weighted clinical loss and conformal residual calibration yield a calibrated upper risk estimate. (3) Joint Ask/Commit search uses this estimate to select questions and determine when to stop. (4) A language interface asks the selected question or reports the leading diagnosis and protected differential. Patient responses update both beliefs before replanning.}
\label{fig:cardi-framework}
\end{figure*}

\section{Related Work}

\paragraph{Interactive medical diagnosis.}
Interactive diagnosis studies evidence gathering through patient consultation~\cite{tchango2022ddxplus,li2024mediq}. Evaluations cover consultations and clinical journeys~\cite{liu2025interactive,fan2025aihospital,wu2024medjourney}, diagnostic reasoning and differential diagnosis~\cite{wang2024direct,rose2025meddxagent}, and the elicitation of necessary evidence~\cite{long-wei-2026-strong}. These settings assess how agents gather evidence, reach a diagnosis, and retain plausible alternatives. Methods for acquiring this evidence include learning question policies with Shapley information gain~\cite{ding-etal-2026-promed} and selecting questions by expected diagnostic information gain~\cite{wong2026medclarify}.

\paragraph{Information acquisition and planning.}
Active information acquisition spans dynamic feature selection, task disambiguation, and Bayesian experimental design~\cite{covert2023dynamic,kobalczyk2025active,choudhury2026bed}. Dialogue policies and clarification strategies guide conversational actions~\cite{deng2024ppdpp,zhang2024clamber,chen2025clarify}, while uncertainty and feedback support planning over question sequences~\cite{hu2024uncertainty,chopra2025feedback,ling2026uncertainty}. Tree search and world models provide mechanisms for exploring future actions~\cite{yao2023tree,hao2023rap,zhou2024lats}. Search mechanisms determine which actions are explored, while the acquisition objective determines how they are valued. Value-of-information methods evaluate questions by their downstream utility relative to interaction cost~\cite{dong-etal-2026-value}. Our work adopts this decision-oriented perspective, evaluating questions through a clinical loss that combines primary diagnosis error with severity-weighted differential omission. It also guides stopping.

\begingroup
\clubpenalties 3 10000 10000 0
\paragraph{Calibration and clinical risk.}
LLM uncertainty research examines elicited confidence~\cite{xiong2024uncertainty}, semantic variation~\cite{kuhn2023semantic,qiu2024semanticdensity}, and calibration of uncertainty estimates~\cite{huang2024rankcalibration,ye2024benchmarking}. Conformal prediction provides finite-sample coverage under exchangeability~\cite{vovk2005algorithmic,romano2020classification}, with extensions that control task losses~\cite{bates2021distribution,angelopoulos2024conformal}. Applications address validity and output selection in language generation~\cite{cherian2024llmvalidity,gui2024conformalalignment}, decisions under risk aversion~\cite{kiyani2025riskaverse}, and medical abstention~\cite{cocchieri-etal-2026-llms}.
\par\endgroup

Calibration and risk control also inform planning with learned dynamics, early classification, and reasoning effort allocation~\cite{sun2023plancp,ringel2024early,wang2026conformalthinking}. Here, calibration targets the estimated clinical loss of committing to a primary diagnosis and differential under partial evidence. Conformal residual calibration corrects risk underestimation to guide both question selection and stopping.

\section{Method}
\label{sec:method}

\subsection{Problem Formulation}
\label{sec:problem}

Given intake information $x_0$, an agent asks follow-up questions to distinguish among diseases in a candidate set $\mathcal{Y}$. After $t$ questions, its history is $h_t=(x_0,(q_1,o_1),\ldots,(q_t,o_t))$, where $q_j$ is a queried feature and $o_j$ its response. An observation mask $M_t$ records queried features. At each turn, \textsc{Ask} requests an unobserved feature, while \textsc{Commit} ends the interaction and returns a leading diagnosis $\hat y_t$ and a protected set $\Gamma_t\subseteq\mathcal{Y}$ containing it and plausible alternatives. With at most $T_{\max}$ questions, the objective is to minimize expected diagnostic loss plus cumulative question costs.

\subsection{Framework Overview}
\label{sec:overview}

CaRDI values evidence by its expected effect on final diagnostic loss and compares this benefit with interaction cost. Figure~\ref{fig:cardi-framework} shows the four components; Algorithm~\ref{alg:cardi} gives the full interaction loop.

\subsection{Clinical Belief Modeling}
\label{sec:belief}

Conditioned on history $h_t$, the diagnostic belief $b_t^D(y)$ estimates primary diagnosis probabilities, while the safety belief $b_t^S(y)$ models support for plausible alternatives. A disease can remain relevant to the differential without being the most likely primary diagnosis, motivating separate beliefs for these two roles.

\paragraph{Belief initialization.}
We estimate disease priors and clinical response likelihoods from smoothed training counts, using primary disease labels for the diagnostic model and normalized weights from the reference differential as fractional disease counts for the safety model. Conditioning these structured models on the intake information $x_0$ yields the posteriors $p_{\mathrm{str}}^D(y\mid x_0)$ and $p_{\mathrm{str}}^S(y\mid x_0)$. The backbone LLM with parameters $\theta$ scores candidate diagnosis names given $x_0$, and softmax normalization yields its initial distribution $p_\theta^0(y)$. We initialize the diagnostic belief by geometrically pooling the structured and LLM distributions:
\begin{equation}
b_0^D(y)
=
\frac{
[p_{\mathrm{str}}^D(y\mid x_0)]^{1-\eta}
[p_\theta^0(y)]^\eta
}{Z_0},
\label{eq:belief-init}
\end{equation}
where $\eta\in[0,1]$ controls the contribution of the LLM and $Z_0$ is the normalization constant. The safety belief is initialized from its structured posterior, $b_0^S(y)=p_{\mathrm{str}}^S(y\mid x_0)$.

\input{sections/algorithm_main}

\paragraph{Bayesian belief updates.}
As new evidence is elicited, we update both beliefs by Bayes' rule. For $r\in\{D,S\}$, let $\ell_r(o\mid q,y)$ denote the likelihood of response $o$ to a query about clinical feature $q$, given disease $y$. Assuming conditional independence of clinical features given the disease, observing $(q_{t+1},o_{t+1})$ yields
\begin{equation}
b_{t+1}^r(y)
=
\frac{
b_t^r(y)\,\ell_r(o_{t+1}\mid q_{t+1},y)
}{
\sum_{y'\in\mathcal{Y}}
b_t^r(y')\,\ell_r(o_{t+1}\mid q_{t+1},y')
}.
\label{eq:belief-update}
\end{equation}
Explicit negative responses contribute evidence; unqueried features add no likelihood factors.

\paragraph{Protected diagnosis set.}
After initialization and each belief update, we construct a protected diagnosis set for commitment. The set retains the leading diagnosis $\hat y_t=\arg\max_{y\in\mathcal{Y}} b_t^D(y)$ together with alternatives supported by the safety belief:
\begin{equation}
\Gamma_t
=
\{\hat y_t\}
\cup
\{y\in\mathcal{Y}:b_t^S(y)>\tau_t\},
\label{eq:protected-set}
\end{equation}
where $\tau_t\in[0,1]$ is an inclusion threshold selected for each turn on a separate held-out subset and fixed before residual calibration.

\subsection{Severity-Aware Conformal Clinical Risk}
\label{sec:conformal-risk}

\paragraph{Clinical Commit loss.}
We define the \emph{Clinical Commit Loss} at turn $t$. Let $Y$ be the reference primary diagnosis and $\lambda\in[0,1]$ balance primary diagnosis error against differential omission:
\begin{equation}
L_t
=
\lambda\,\mathbf{1}\{\hat y_t\neq Y\}
+
(1-\lambda)\Omega_t,
\label{eq:commit-loss}
\end{equation}
where the indicator records primary diagnosis error and $\Omega_t$ measures differential omission. Let $g(y)$ be the normalized reference-differential weight for disease $y$ and $\kappa(y)\geq1$ its prespecified severity weight; $g(y)=0$ outside the annotated diagnoses. The weighted omitted fraction is
\begin{equation}
\Omega_t
=
\frac{
\sum_{y\notin\Gamma_t}\kappa(y)g(y)
}{
\sum_{y\in\mathcal{Y}}\kappa(y)g(y)
}.
\label{eq:true-omission}
\end{equation}
Larger $\kappa(y)$ penalizes severe omissions more heavily; normalization keeps the loss in $[0,1]$. Retaining an annotated severe alternative can therefore lower omission loss even when the leading diagnosis remains unchanged, giving information acquisition a second route to improving the final output.

\paragraph{Risk estimation.}
During interaction, reference diagnoses are unavailable. We estimate Commit risk $R_t$ using $1-b_t^D(\hat y_t)$ for diagnosis error and $b_t^S(y)$ in place of $g(y)$ for omission:
\begin{equation}
\begin{aligned}
R_t ={}& \lambda\bigl(1-b_t^D(\hat y_t)\bigr) \\
&+ (1-\lambda)
\frac{\sum_{y\notin\Gamma_t}\kappa(y)b_t^S(y)}
     {\sum_{y\in\mathcal{Y}}\kappa(y)b_t^S(y)}.
\end{aligned}
\label{eq:plugin-risk}
\end{equation}
A question can lower $R_t$ by improving primary diagnosis or clarifying which alternatives to retain.

\paragraph{Conformal calibration.}
We correct for risk underestimation using one-sided split conformal calibration~\cite{vovk2005algorithmic,bates2021distribution,angelopoulos2024conformal}. At each turn $t$, the residual $E_i^{(t)}=L_i^{(t)}-R_i^{(t)}$ measures the estimation error on held-out case $i$. Positive residuals indicate underestimated clinical loss. We obtain a correction $\widehat q_t$ from the conformal quantile of these residuals at target miscoverage level $\alpha$. The calibrated upper risk estimate is
\begin{equation}
U_t
=
\min\!\left\{
1,\,
R_t+\max(0,\widehat q_t)
\right\}.
\label{eq:upper-risk}
\end{equation}
The residual connects realized diagnostic consequences to the belief-based estimate used during interaction. Taking the positive part of the residual quantile adds a nonnegative margin, while clipping at one keeps the estimate within the range of the clinical loss.

At a fixed turn $t$, exchangeability of the calibration residuals and a fresh test residual gives
\begin{equation}
\Pr\!\left(
L_t^{\mathrm{test}}\leq U_t^{\mathrm{test}}
\right)
\geq1-\alpha.
\label{eq:coverage}
\end{equation}
The probability is marginal over the calibration sample and the fresh test case. Appendix~\ref{app:conformal} gives the quantile construction, proof, and \mbox{state collection protocol}.

\subsection{Risk-Guided Planning}
\label{sec:planning}

Given the history, mask, and both beliefs in state $s_t$, the planner compares immediate commitment at risk $U(s_t)=U_t$ with further questioning.

\paragraph{Questions and responses.}
Each belief proposes unqueried clinical features linked to its high-probability diseases. The shortlist $\mathcal{Q}^{\mathrm{cand}}(s)$ reserves slots for both beliefs and keeps at most $K_q$ questions ranked by expected reduction in $U$.

To simulate the outcome of asking $q$ at state $s$, we compute response probabilities by averaging disease-specific likelihoods under the diagnostic belief $b_s^D(y)$:
\begin{equation}
p(o\mid s,q)
=
\sum_{y\in\mathcal{Y}}b_s^D(y)\,\ell_D(o\mid q,y).
\label{eq:response-model}
\end{equation}
Each simulated response induces a transition $s'=T(s,q,o)$, updating the history, observation mask, and both beliefs via Equation~\eqref{eq:belief-update}, then recomputing the protected set and risk. A candidate question may lead to early commitment after one response and further questioning after another. Evaluating these branches accounts for how the same question changes both the diagnosis and \mbox{the remaining interview}.

\paragraph{Planning objective.}
Let $c(q)\geq0$ denote the cost of asking $q$, measured on the same scale as Commit loss. Consider a simulated trajectory $\xi$ beginning at $s_0=s$, with $m$ questions $q_0,\ldots,q_{m-1}$ and terminal state $s_m$. Its return is
\begin{equation}
G(\xi)
=
-\sum_{j=0}^{m-1}c(q_j)-U(s_m),
\label{eq:trajectory-return}
\end{equation}
where $j$ indexes questions within the simulated trajectory. Immediate commitment has $m=0$ and return $-U(s)$. Search maximizes expected return by evaluating how responses affect subsequent questions and stopping decisions. Further questioning is preferred when its expected reduction in terminal risk outweighs its cumulative cost. The correction for turn $t+m$ enters $U(s_m)$, allowing calibration to affect the value of future questioning through its possible stopping points.

\paragraph{Search and action selection.}
Monte Carlo Tree Search (MCTS)~\cite{silver2010monte,chopra2025feedback} estimates action values from simulated returns, using expected immediate risk reduction as an action prior. Search uses $B$ simulations with depth limited to $H$ or the remaining question budget, whichever is smaller. Commit is available at every node with value $-U(s)$. States at the search depth limit receive the same terminal value. MCTS returns the highest-valued root action $a_t^\star$. Appendix~\ref{app:mcts} describes the search.

\subsection{Language-Grounded Action Execution}
\label{sec:execution}

The language interface maps the selected action to a patient-facing question or diagnostic report. An Ask response updates the history, mask, and beliefs before replanning (Section~\ref{sec:belief}); Commit returns the leading diagnosis and protected set. In medical question answering, responses trigger LLM rescoring, and Commit returns the most probable option. Appendix~\ref{app:execution} \mbox{details the interface}.

\input{DDXPlus_ACL_main_table_20260917.tex}

\begin{table*}[!t]
  \centering
  \small
  \setlength{\tabcolsep}{2.5pt}
  \renewcommand{\arraystretch}{1.08}
  \begin{tabularx}{\textwidth}{@{}lc*{5}{>{\centering\arraybackslash}X}c@{}}
    \toprule
    Method & ECCL$\downarrow$ & \mbox{Sev. Omit.$\downarrow$} & Top-1$\uparrow$ & DDx-F1$\uparrow$ & Sev-F1$\uparrow$ & Queries$\downarrow$ & Risk Exc.$\downarrow$ \\
    \midrule
    w/o Safety Belief & 0.177 & 49.39 & 95.44 & 0.320 & 0.210 & \underline{3.41} & 36.80 \\
    w/o Risk-Guided Planning & 0.075 & 38.98 & \textbf{96.55} & \textbf{0.629} & \textbf{0.576} & 3.95 & 9.32 \\
    w/o Conformal Calibration & 0.073 & 35.96 & 96.08 & 0.618 & 0.562 & \textbf{3.32} & 37.35 \\
    w/o Joint Ask/Commit & \underline{0.069} & \underline{33.94} & 95.94 & 0.616 & 0.559 & 3.56 & \underline{7.14} \\
    \rowcolor{cardigreen}
    \textbf{Full CaRDI} & \textbf{0.059} & \textbf{29.86} & \underline{96.37} & \underline{0.623} & \underline{0.571} & 3.58 & \textbf{6.87} \\
    \bottomrule
  \end{tabularx}
  \caption{Ablation study of CaRDI on DDXPlus. Sev. Omit. measures omission of severe diseases in the reference differential; Risk Exc. measures risk exceedance at commitment.}
  \label{tab:ablation-study}
\end{table*}

\section{DDXPlus: Diagnosis and Differential Retention}

\subsection{Experimental Setup}
\paragraph{Dataset.}
DDXPlus~\cite{tchango2022ddxplus} provides patient records with clinical findings and reference differentials. We evaluate on 490 cases from its official test split. The intake gives age, sex, and one positive finding; questions then reveal recorded symptoms or antecedents.

\paragraph{Baselines.}
We compare an intake-only reference with six active methods:
\begin{itemize}
\renewcommand{\labelitemi}{$\bullet$}
\setlength{\itemsep}{2pt}
\setlength{\parsep}{0pt}
\setlength{\topsep}{3pt}
\item \textbf{Non-interactive diagnosis.} Direct Commit predicts from the intake without questions.
\item \textbf{Active acquisition.} ATD~\cite{kobalczyk2025active} asks questions that distinguish diagnostic hypotheses. UoT~\cite{hu2024uncertainty} and MISQ-HF~\cite{chopra2025feedback} use uncertainty-aware planning. C-IP~\cite{chan2025conformal} ranks questions with a conformal prediction-set entropy bound. ACTMED~\cite{estevez2025actmed} and BED-LLM~\cite{choudhury2026bed} use expected multi-disease KL divergence and information gain, respectively.
\end{itemize}

\paragraph{Evaluation metrics.}
We report Top-1/Top-3 accuracy and mean additional questions (Queries). Empirical Clinical Commit Loss (ECCL) evaluates final diagnostic loss, while DDx-F1 measures agreement with the reference differential. Severity-F1 summarizes severe-disease rule-in and rule-out; Severe Error measures Top-1 error on severe cases. Top-1, Top-3, and Severe Error are reported as percentages; ECCL and F1 scores range from 0 to 1. Main results use each method's returned differential. In Tables~\ref{tab:overall-performance} and~\ref{tab:ablation-study}, we bold the best values and underline the second best.

\paragraph{Implementation details.}
We use Qwen3-4B \citep{qwenteam2025qwen3}, Mistral-7B-Instruct-v0.3 \citep{mistralai2024mistral}, and Llama-3.1-8B-Instruct \citep{llamateam2024llama3}. The question budget is ten, except for ATD's four-question limit. ECCL weights diagnostic error and omission equally, with a twofold cost for severe omission. 

Appendix~\ref{app:ddx-setup} details data splits, baseline adaptations, search settings, and metric definitions.

\begin{table}[t]
\centering
\fontsize{10}{12}\selectfont
\setlength{\tabcolsep}{3pt}
\renewcommand{\arraystretch}{1.12}
\begin{tabularx}{\columnwidth}{Xrrr}
\toprule
Case group & $n$ & Acc. gain (pp) & ECCL reduction \\
\midrule
All cases & 980 & 2.99 & 0.0187 \\
Low entropy & 352 & 1.52 & 0.0095 \\
\rowcolor{heatblue!8}
High entropy & 338 & \textbf{6.80} & \textbf{0.0445} \\
\bottomrule
\end{tabularx}
\caption{Gains from selecting the first question by clinical risk rather than entropy on DDXPlus.}
\label{tab:analysis-first-gains}
\end{table}

\subsection{Main Results}
\label{sec:ddx-main-results}
\begin{figure}[t]
\centering
\includegraphics[width=\columnwidth]{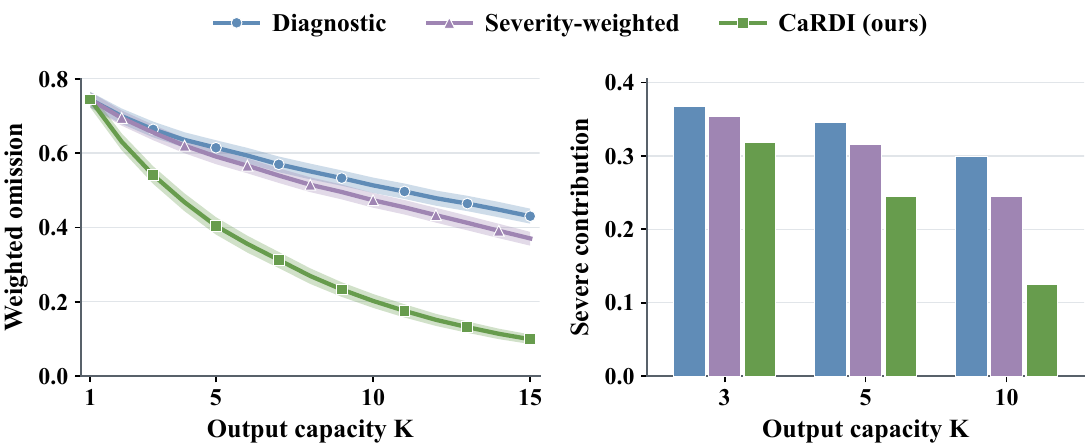}
\caption{Differential retention across output capacities on DDXPlus. Left: weighted omission; right: the contribution from severe diseases. Bands show 95\% patient-bootstrap confidence intervals.}
\label{fig:output-capacity}
\end{figure}

\begingroup
\clubpenalties 5 10000 10000 10000 10000 0
Table~\ref{tab:overall-performance} summarizes performance across three backbones, covering primary diagnosis, severe-disease recognition, and differential retention alongside the number of questions asked.

\paragraph{Higher accuracy with fewer questions.}
CaRDI achieves the highest Top-1 accuracy across all three backbones. Compared with BED-LLM, the strongest baseline on this metric, it gains about nine percentage points while asking 10--13\% fewer questions. All active methods outperform Direct Commit, but more questions do not consistently yield greater accuracy.

\paragraph{Better recognition of severe diseases.}
CaRDI reduces severe-case error to at most 4\%, compared with 8.7\% for BED-LLM, and achieves the highest Severity-F1. These gains improve both primary diagnosis and recognition of severe alternatives.

\paragraph{Lower loss and better differentials.}
CaRDI achieves the lowest ECCL and highest DDx-F1 across backbones. CaRDI and ACTMED return an average of 15.2 and 3.8 diagnoses, respectively. The larger CaRDI differential reduces \mbox{omission loss} and accounts for much of the ECCL advantage. At a common output size of three diagnoses, ECCL is about 13\% lower than ACTMED's across backbones, driven by more accurate primary diagnoses. Figure~\ref{fig:output-capacity} isolates retention quality at fixed output capacity.
\par\endgroup

\subsection{Ablation Study}
\label{sec:ablation-study}

Table~\ref{tab:ablation-study} summarizes the component ablations.

\paragraph{Risk-guided planning.}
We replace clinical risk with predictive entropy while keeping the belief and response models and search settings fixed. Using the risk objective lowers ECCL by 0.016 and reduces severe omission with fewer questions.

\paragraph{Safety belief.}
Removing the safety belief triples ECCL while leaving diagnostic accuracy high, highlighting the importance of preserving alternatives beyond the leading diagnosis.

\paragraph{Calibration and joint planning.}
Without conformal calibration, CaRDI commits earlier and incurs higher ECCL. Removing joint Ask/Commit planning also raises loss at nearly the same query count, supporting joint Ask/Commit decisions.

\begin{table}[t]
    \centering
    \small
    \setlength{\tabcolsep}{3pt}
    \renewcommand{\arraystretch}{1.08}
    \begin{tabularx}{\columnwidth}{@{}l*{4}{>{\centering\arraybackslash}X}}
        \toprule
        Backbone & $K=3$ & $K=5$ & $K=10$ & $K=15$ \\
        \midrule
        \backbonelogo{0.18in}{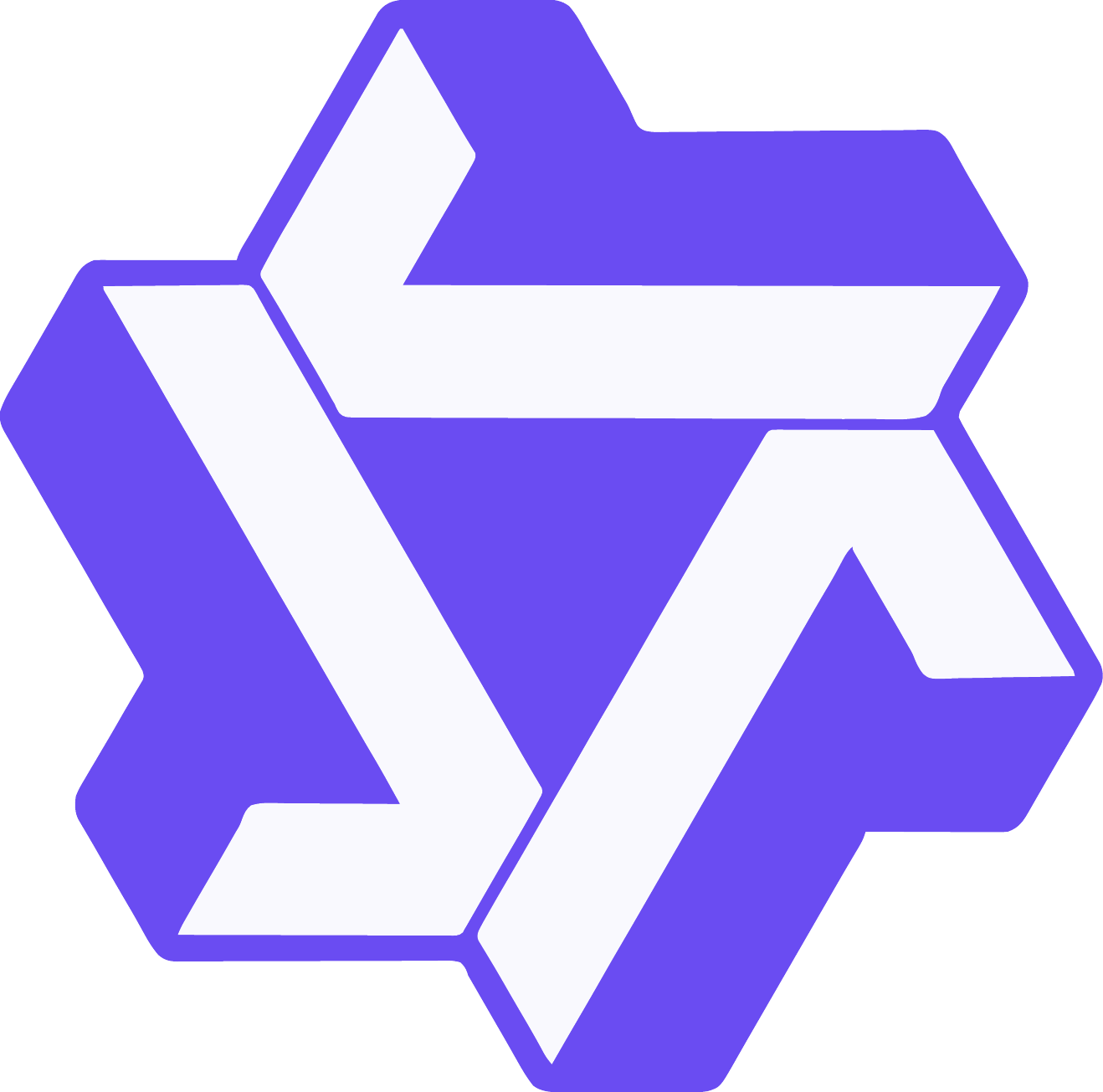}\,Qwen3-4B
        & \heatcell{16}{18.2}
        & \heatcell{34}{31.8}
        & \heatcell{70}{58.1}
        & \heatcell{91}{\textbf{73.6}} \\

        \backbonelogo{0.18in}{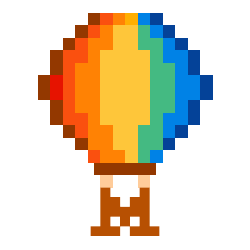}\,Mistral-7B
        & \heatcell{14}{16.6}
        & \heatcell{33}{31.1}
        & \heatcell{70}{58.0}
        & \heatcell{92}{\textbf{74.2}} \\

        \backbonelogo{0.18in}{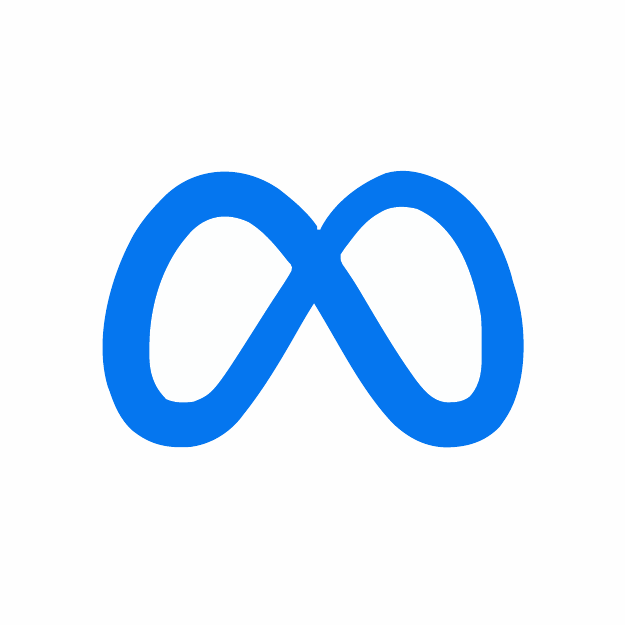}\,Llama-3.1-8B
        & \heatcell{15}{17.4}
        & \heatcell{34}{31.6}
        & \heatcell{69}{57.2}
        & \heatcell{90}{\textbf{73.2}} \\
        \bottomrule
    \end{tabularx}
    \caption{Relative reduction (\%) in weighted omission by CaRDI versus severity-weighted diagnostic ranking. $K$ denotes the number of retained diagnoses; darker shading indicates larger reductions.}
    \label{tab:omission_reduction}
\end{table}

\input{sections/mediq_figure}

\begingroup
\clubpenalties 3 10000 10000 0
\input{sections/analysis}
\par\endgroup

\input{sections/mediq}

\section{Conclusion}
CaRDI aligns information acquisition and stopping with calibrated final diagnostic loss. Experiments on DDXPlus and MediQ show improved accuracy with efficient questioning. Controlled analyses establish gains in first-question quality and differential retention, and show how calibration affects planning through stopping. These findings support valuing questions and stopping decisions by their effects on diagnostic accuracy and the retention of severe alternatives in the final differential.

\section*{Limitations}
DDXPlus tests structured finding likelihoods with an LLM intake prior; MediQ tests language-based elicitation under an answer-error objective. Both benchmarks use English, with recorded or simulated responses. Benchmark-derived severity weights approximate clinical consequences. Equation~\eqref{eq:coverage} requires residual exchangeability at a fixed turn, while random-query calibration and adaptive deployment can induce different residual distributions. Extending the evaluation to free-form patient dialogue requires testing patient-specific consequence estimates and distribution shift.

\section*{Ethical Considerations}
CaRDI is intended for research in diagnostic decision support. Clinical use would require validation, clinician oversight, privacy protection, and evaluation of diagnostic errors \mbox{across patient groups}.

\section*{Acknowledgements}
The work described in this paper was partially supported by InnoHK initiative, The Government of the HKSAR, and Laboratory for AI-Powered Financial Technologies.

\bibliography{references}

\clearpage
\appendix
\newtcolorbox{casebox}[1]{%
  colback=white,colframe=black!30,colbacktitle=black!6,coltitle=black,
  title={#1},fonttitle=\small\bfseries,fontupper=\small\rmfamily,
  boxrule=0.4pt,arc=1pt,left=7pt,right=7pt,top=6pt,bottom=6pt,
  before skip=8pt,after skip=8pt,
  before upper={\raggedright\setlength{\parskip}{4pt}},
  toptitle=3pt,bottomtitle=3pt
}
\makeatletter
\setlength{\@dblfptop}{0pt}
\makeatother

\input{sections/algorithm_appendix}

\input{sections/calibration_appendix}

\input{sections/ddx_appendix}
\input{sections/analysis_appendix}
\FloatBarrier
\input{sections/mediq_protocol}

\end{document}

%% file: sections/intro_empirical.tex
\section{Introduction}
Medical consultation gathers patient evidence through targeted questions and commits to a diagnosis once the evidence suffices. Recent advances in large language models (LLMs) have extended automated diagnostic support from answering medical questions to conducting such interactive consultations~\cite{singhal2023clinical,li2024mediq,ding-etal-2026-promed}. In this setting, an LLM agent must reason with incomplete evidence while deciding which information to acquire next and when to conclude the consultation. Its questions shape the evidence available for diagnosis, while prolonged interaction imposes additional cost. Effective consultation therefore requires coordinating evidence acquisition with the diagnostic \mbox{decisions it supports}.

\begin{figure}[t]
\centering
\includegraphics[width=\columnwidth,trim=3bp 3bp 3bp 3bp,clip]{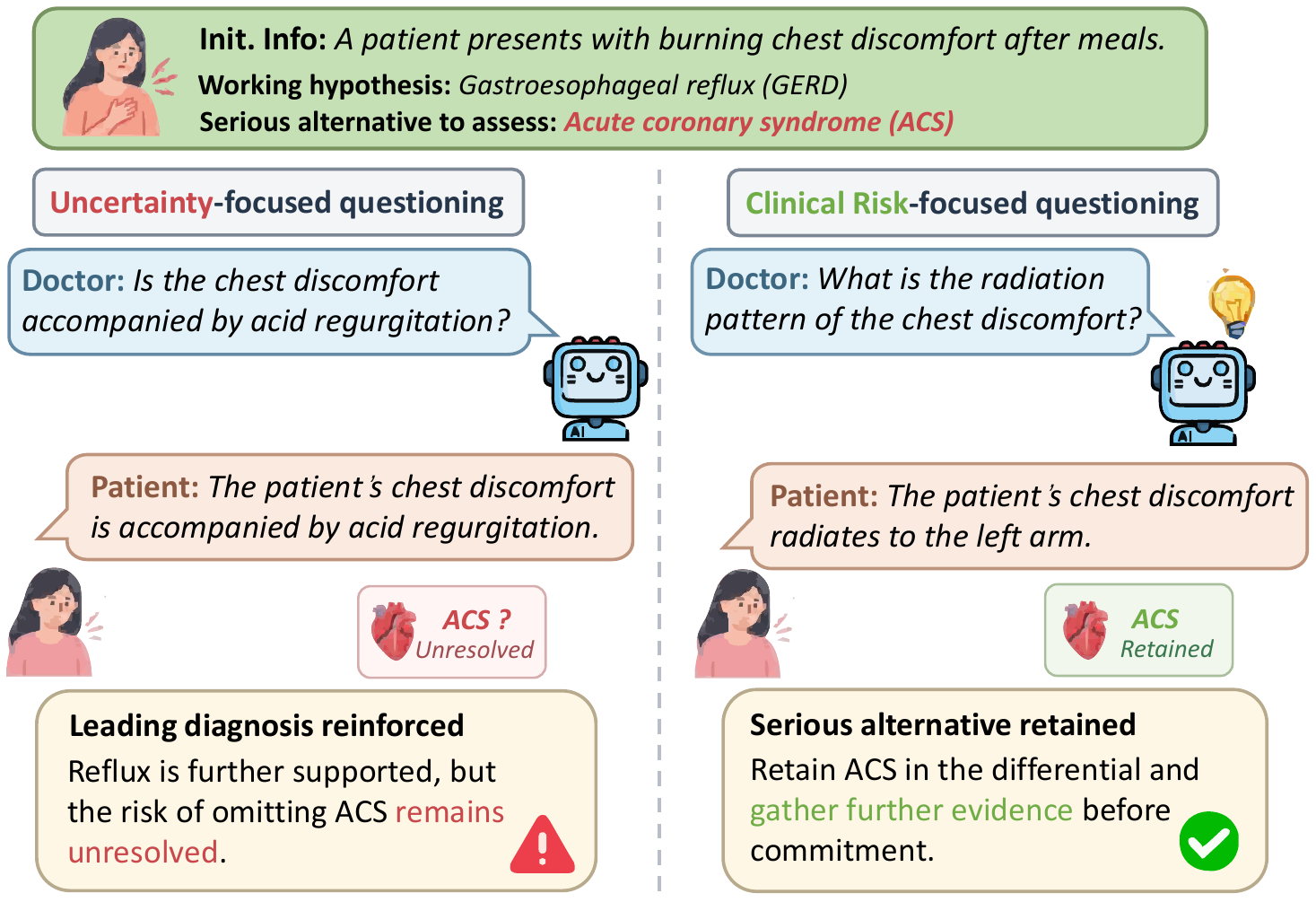}
\caption{Illustration of question selection in interactive diagnosis. Uncertainty-focused questions reinforce reflux, while risk-focused questions elicit evidence for retaining acute coronary syndrome in~the~differential.}
\label{fig:intro-motivation}
\end{figure}

Building on LLMs, recent efforts in interactive diagnosis and information seeking address both when to request additional evidence and which questions to ask, mostly by quantifying uncertainty about the diagnosis. MediQ uses model confidence to defer diagnostic decisions and elicit missing patient information~\cite{li2024mediq}. To select informative questions, BED-LLM applies Bayesian experimental design, while MedClarify uses expected information gain to distinguish among candidate diagnoses~\cite{choudhury2026bed,wong2026medclarify}. C-IP estimates uncertainty with conformal prediction sets, and UoT evaluates possible future responses to guide question selection~\cite{chan2025conformal,hu2024uncertainty}. A complementary approach values questions by their expected improvement in downstream decisions relative to interaction cost~\cite{dong-etal-2026-value}. These approaches differ in what makes a question valuable: reducing ambiguity or improving the decision supported by \mbox{the acquired evidence}.

However, uncertainty reduction alone does not distinguish the clinical consequences of primary diagnosis errors and omitted severe alternatives. The same predictive uncertainty can therefore correspond to different clinical risks when unresolved alternatives differ in severity. As Figure~\ref{fig:intro-motivation} illustrates, a question that strengthens a reflux hypothesis may increase diagnostic confidence without resolving whether acute coronary syndrome should remain in the differential. This reveals an \textbf{objective mismatch between uncertainty reduction and final diagnostic loss}. Addressing this mismatch requires valuing questions by their expected effect on primary diagnosis error and severe omissions, with both risks estimated from \mbox{incomplete patient evidence}.

To address this challenge, we propose \emph{Calibrated Risk for Diagnostic Interaction} (\textbf{CaRDI}), a planning framework that uses final diagnostic loss as a common objective for evidence acquisition and stopping. We define a \emph{Clinical Commit Loss} that combines primary diagnosis error with severity-weighted differential omission. Diagnostic and safety beliefs estimate these complementary components, while conformal residual calibration adjusts for underestimation using held-out cases. The resulting calibrated risk guides whether and what to ask next by balancing the expected reduction in diagnostic loss against interaction cost. Experiments on DDXPlus and MediQ across multiple LLM backbones show higher diagnostic accuracy with fewer questions than uncertainty-driven baselines. Further analysis on DDXPlus shows that replacing entropy with clinical risk as the planning objective reduces clinical loss by 21.5\% and decreases severe omissions.

Our main contributions are as follows:
\begin{itemize}
\setlength{\itemsep}{2pt}
\setlength{\parskip}{0pt}
\item[$\bullet$] We formulate decision-risk planning for interactive diagnosis using a clinical loss that combines primary diagnosis error with severity-weighted differential omission.
\item[$\bullet$] We develop CaRDI, integrating complementary beliefs and conformal residual calibration to guide question selection and stopping with a shared estimate of clinical risk.
\begin{samepage}
\item[$\bullet$] We demonstrate gains across benchmarks and backbones, with controlled comparisons showing improved diagnostic outcomes at comparable mean post-response entropy.
\end{samepage}
\end{itemize}

%% file: sections/algorithm_main.tex
\begin{algorithm}[t]
\caption{CaRDI for Interactive Diagnosis}
\label{alg:cardi}
\small
\renewcommand{\algorithmicrequire}{\textbf{Input:}}
\begin{algorithmic}[1]
\REQUIRE Intake $x_0$; fitted models and calibration;\\
cost $c$, budget $T_{\max}$, depth $H$, simulations $B$
\STATE $s_0\gets\operatorname{Initialize}(x_0)$
\FOR{$t=0,\ldots,T_{\max}$}
    \STATE $\hat y_t\gets\arg\max_y b_t^D(y)$
    \STATE $\Gamma_t\gets\{\hat y_t\}\cup\{y:b_t^S(y)>\tau_t\}$
    \STATE $U(s_t)\gets\operatorname{CalibratedRisk}(s_t,\Gamma_t,\widehat q_t)$
    \STATE $\mathcal Q_t\gets\mathcal Q^{\mathrm{cand}}(s_t)$
    \STATE \textbf{if} $t=T_{\max}$ or $\mathcal Q_t=\varnothing$ \textbf{then break}
    \STATE $H_t\gets\min(H,T_{\max}-t)$
    \STATE $a_t^\star\gets\operatorname{MCTS}(s_t,\mathcal Q_t,U,c,H_t,B)$
    \STATE \textbf{if} $a_t^\star=\operatorname{Commit}$ \textbf{then break}
    \STATE Let $a_t^\star=\operatorname{Ask}(q_{t+1}^\star)$
    \STATE Ask $q_{t+1}^\star$ and observe $o_{t+1}$
    \STATE $s_{t+1}\gets T(s_t,q_{t+1}^\star,o_{t+1})$
\ENDFOR
\STATE Commit and \textbf{return} $\hat y_t,\Gamma_t$
\end{algorithmic}
\end{algorithm}

%% file: DDXPlus_ACL_main_table_20260917.tex
\begin{table*}[t]
\centering
\small
\setlength{\tabcolsep}{3pt}
\renewcommand{\arraystretch}{1.08}
\begin{tabularx}{\textwidth}{@{}ll*{5}{>{\centering\arraybackslash}X}cc@{}}
\toprule
Model & Method & Top-1$\uparrow$ & Top-3$\uparrow$ & Queries$\downarrow$ & ECCL$\downarrow$ & DDx-F1$\uparrow$ & Sev-F1$\uparrow$ & Sev. Err.$\downarrow$ \\
\midrule
\multirow{9}{*}{\emph{Qwen3-4B}} & Direct & 32.86 & 56.73 & 0.00 & 0.543 & 0.391 & 0.249 & 66.96 \\
 & ATD & 61.22 & 87.14 & \underline{4.00} & 0.485 & 0.346 & \underline{0.180} & 32.17 \\
 & UoT & 65.10 & 90.61 & 5.42 & 0.478 & 0.330 & 0.154 & 29.57 \\
 & MISQ-HF & 66.53 & 88.16 & 6.73 & 0.474 & 0.333 & 0.155 & 33.04 \\
 & C-IP & 72.45 & 84.08 & 4.03 & 0.432 & 0.341 & 0.169 & 26.09 \\
 & ACTMED & 85.10 & \underline{91.63} & 6.22 & \underline{0.396} & \underline{0.360} & 0.124 & 18.26 \\
 & BED-LLM & \underline{87.35} & 89.18 & 4.04 & 0.442 & 0.279 & 0.080 & \underline{8.70} \\
\rowcolor{cardigreen}
 & \textbf{CaRDI (Ours)} & \textbf{96.12} & \textbf{99.31} & \textbf{3.64} & \textbf{0.060} & \textbf{0.625} & \textbf{0.572} & \textbf{4.00} \\
\rowcolor{heatblue!10}
 & \textit{vs. baseline} & +8.77 & +7.68 & +0.36 & +0.336 & +0.265 & +0.392 & +4.70 \\
\midrule
\multirow{9}{*}{\emph{Mistral-7B}} & Direct & 32.86 & 56.53 & 0.00 & 0.540 & 0.393 & 0.250 & 66.96 \\
 & ATD & 61.63 & 87.35 & 4.00 & 0.481 & 0.352 & \underline{0.180} & 30.43 \\
 & UoT & 64.49 & 90.82 & 5.42 & 0.481 & 0.330 & 0.154 & 30.43 \\
 & MISQ-HF & 67.55 & 89.59 & 7.47 & 0.463 & 0.345 & 0.163 & 24.35 \\
 & C-IP & 73.67 & 82.24 & \underline{3.75} & 0.431 & 0.328 & 0.147 & 26.96 \\
 & ACTMED & 85.51 & \underline{92.04} & 6.33 & \underline{0.395} & \underline{0.359} & 0.124 & 20.00 \\
 & BED-LLM & \underline{87.35} & 89.18 & 4.05 & 0.442 & 0.279 & 0.080 & \underline{8.70} \\
\rowcolor{cardigreen}
 & \textbf{CaRDI (Ours)} & \textbf{96.69} & \textbf{99.35} & \textbf{3.54} & \textbf{0.058} & \textbf{0.626} & \textbf{0.574} & \textbf{3.13} \\
\rowcolor{heatblue!10}
 & \textit{vs. baseline} & +9.34 & +7.31 & +0.21 & +0.337 & +0.267 & +0.394 & +5.57 \\
\midrule
\multirow{9}{*}{\emph{Llama-3.1-8B}} & Direct & 33.06 & 56.53 & 0.00 & 0.538 & 0.393 & 0.249 & 66.96 \\
 & ATD & 63.06 & 86.73 & \underline{4.00} & 0.476 & 0.346 & \underline{0.177} & 29.57 \\
 & UoT & 64.69 & 90.82 & 5.42 & 0.480 & 0.330 & 0.154 & 30.43 \\
 & MISQ-HF & 66.73 & 87.14 & 7.71 & 0.470 & 0.334 & 0.150 & 33.04 \\
 & C-IP & 72.04 & 81.84 & 4.09 & 0.435 & 0.335 & 0.163 & 26.09 \\
 & ACTMED & 84.90 & \underline{91.63} & 6.24 & \underline{0.397} & \underline{0.361} & 0.124 & 20.87 \\
 & BED-LLM & \underline{87.35} & 89.18 & 4.04 & 0.442 & 0.279 & 0.080 & \underline{8.70} \\
\rowcolor{cardigreen}
 & \textbf{CaRDI (Ours)} & \textbf{96.37} & \textbf{99.35} & \textbf{3.58} & \textbf{0.059} & \textbf{0.623} & \textbf{0.571} & \textbf{3.30} \\
\rowcolor{heatblue!10}
 & \textit{vs. baseline} & +9.02 & +7.72 & +0.42 & +0.338 & +0.262 & +0.394 & +5.40 \\
\bottomrule
\end{tabularx}
\caption{Overall performance on DDXPlus across three backbones. Best results are bolded, and second-best results are underlined. Gains are measured against the best active baseline for each metric. }
\label{tab:overall-performance}
\end{table*}

%% file: sections/mediq_figure.tex
\begin{figure*}[t]
\centering
\includegraphics[width=\textwidth]{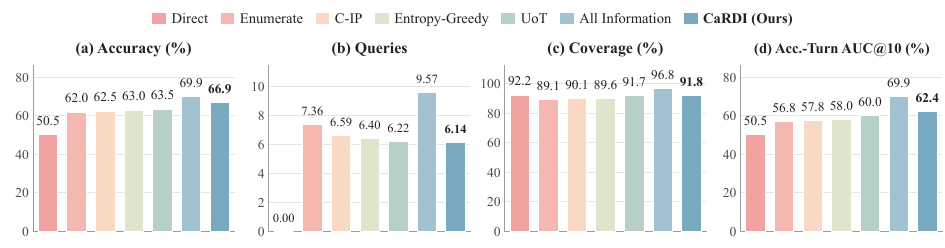}
\caption{Performance on MediQ. CaRDI achieves the highest accuracy among methods using partial information, with about six questions. Coverage measures inclusion of the reference option at commitment; AUC@10 summarizes accuracy over turns 0--10.}
\label{fig:mediq-overall}
\end{figure*}

%% file: sections/analysis.tex
\subsection{Analysis}
\label{sec:analysis}

\paragraph{Better first-question outcomes.}
With Llama-3.1-8B, one response, and five retained diagnoses, risk-guided planning improves accuracy by 2.99 percentage points and reduces ECCL by 0.0187 across 980 cases (Table~\ref{tab:analysis-first-gains}). Mean normalized post-response entropies differ by only 0.001, showing that similar average uncertainty can accompany different diagnostic outcomes. Gains vary across initial-entropy groups, with the largest observed gains in the high-entropy group: 6.80 percentage points higher accuracy and 0.0445 lower ECCL. The fixed response and output budgets focus this comparison on the value of the initial query, complementing the end-to-end results that also involve stopping and longer question sequences.

\paragraph{Safety beliefs improve retention.}
Figure~\ref{fig:output-capacity} compares retention rules using the same terminal evidence, primary predictions, and output capacity. Each rule retains the leading diagnosis and fills the remaining slots with its preferred alternatives. The comparison measures how well these slots preserve severity-weighted reference mass, linking retention quality to the omission term in the clinical loss. With five diagnoses per output, CaRDI reduces weighted omission by 34.2\% relative to diagnostic ranking on Llama. The reduction is about 31\% across backbones when compared with severity-weighted diagnostic ranking. Table~\ref{tab:omission_reduction} reports the corresponding reductions across backbones and capacities. The safety belief therefore improves differential retention beyond severity reweighting alone. Appendix~\ref{app:output-capacity} gives the full capacity curves and paired intervals.

\paragraph{Severity weights protect severe alternatives.}
With beliefs and planning resources fixed, we compare twofold severe-disease costs with uniform costs. Evaluated under the same cost function, severity weighting reduces the severe-disease contribution to weighted omission by 8.5\%. Both variants achieve 96.4\% accuracy with about 3.6 questions. The benefit therefore lies in preserving severe alternatives in the final differential.

\paragraph{Calibration guides questions through stopping.}
We replay recorded decisions with beliefs and search settings fixed. In exact two-step search, removing the correction changes the selected action at 17 of 205 states preceding the final question. Branches can commit at different turns with different risk corrections, altering the relative value of candidate questions. Calibration thus influences question selection through future stopping decisions. Appendix~\ref{app:analysis} gives the full replay results.

%% file: sections/mediq.tex
\section{Extension to Interactive Medical Question Answering}
\label{sec:mediq}
We also evaluate interactive medical question answering on MediQ, examining how calibrated risk guides question selection and stopping as patient responses inform the final answer.

\subsection{Task and Evaluation Setup}
\textbf{Task and baselines.} MediQ~\cite{li2024mediq} covers four-option medical question answering in internal medicine, pediatrics, and neurology. Given a partial vignette, the Expert selects cached questions and rescores options after each Patient response. Baselines include Direct, Enumerate, Entropy-Greedy, UoT~\cite{hu2024uncertainty}, and \mbox{C-IP}~\cite{chan2025conformal}. All Information uses~the~full~record.

\noindent\textbf{Implementation details.} We use residual calibration at $\alpha=0.1$, a ten-question budget, and \mbox{Llama-3.1-8B-Instruct}. For MediQ, we set $\lambda=1$ so that commitment loss reduces to answer error. This specialization retains the Ask/Commit tradeoff with a choice among four answer options. Each response can change the option scores, so its value depends on the final answer it supports. MCTS evaluates candidate questions through empirical transitions between belief \mbox{summaries}, and the Expert replans after each response. Appendix~\ref{app:mediq-protocol} gives implementation and metric details.

\subsection{Accuracy and Interaction Efficiency}
\paragraph{Higher accuracy at similar cost.}
Figure~\ref{fig:mediq-overall} shows that CaRDI achieves the highest accuracy among partial-information methods. It improves on UoT from 63.5\% to 66.9\%, with both methods asking about six questions. This gain closes more than half of UoT's gap to All Information.

\paragraph{Accuracy throughout the interaction.}
CaRDI also achieves the highest accuracy--turn AUC among partial-information methods, indicating better average accuracy over the interview. Its answer set retains the reference option in 91.8\% of cases at commitment.

%% file: sections/algorithm_appendix.tex
\section{Implementation Details}
\label{app:algorithm}
The following details support Algorithm~\ref{alg:cardi}: belief fitting and initialization, candidate generation, search, and action execution. Belief models, thresholds, and residual corrections are fitted before interaction.

\subsection{Belief Models and Initialization}
\label{app:backbone}

\paragraph{Structured likelihoods.}
The two models use the same training cases but different label weights. The diagnostic model counts finding outcomes under primary-disease labels. The safety model uses normalized reference-differential weights as fractional disease counts. Smoothed counts define disease priors and categorical response likelihoods, including the absent outcome. At interaction time, the likelihood product uses revealed findings.

\paragraph{Label scoring.}
The frozen backbone receives the intake information and the list of 49 candidate diagnoses. For the token sequence $(z_{y,1},\ldots,z_{y,m_y})$ of diagnosis $y$, we compute its mean conditional token log-likelihood,
\begin{equation*}
s_\theta(y;x_0)=\frac{1}{m_y}\sum_{j=1}^{m_y}
\log p_\theta(z_{y,j}\mid x_0,z_{y,<j}),
\end{equation*}
where $x_0$ in the conditioning denotes the formatted intake prompt, including the candidate list. Normalization over the fixed label space gives
\begin{equation*}
p_\theta^0(y)=\frac{\exp s_\theta(y;x_0)}{\sum_{y'\in\mathcal{Y}}\exp s_\theta(y';x_0)}.
\end{equation*}
Scoring uses the intake and observed finding.

\paragraph{Belief updates.}
The intake score is cached for the episode. Structured beliefs incorporate each observed or simulated response via Equation~\eqref{eq:belief-update}.

\subsection{Joint Ask/Commit Search}
\label{app:mcts}

\paragraph{Clinical-risk prior.}
The action prior favors questions with a larger expected immediate reduction in upper risk. For a feasible question $q$, the unnormalized Ask weight is
\begin{equation*}
w(s,\operatorname{Ask}(q))
=
\max\!\left\{
U(s)-\mathbb{E}_{o\mid s,q}[U(s')],\,
\varepsilon
\right\},
\end{equation*}
where $s'=T(s,q,o)$ and responses follow Equation~\eqref{eq:response-model}. The floor $\varepsilon>0$ ensures a positive weight for every candidate question. Commit receives the mean Ask weight, or weight one if no Ask action is feasible. Let $\mathcal{A}(s)$ be the feasible actions. The normalized prior is
\begin{equation}
P(s,a)
=
\frac{w(s,a)}{\sum_{a'\in\mathcal{A}(s)}w(s,a')}.
\label{eq:risk-prior}
\end{equation}

\paragraph{Candidate actions and response branches.}
Each belief proposes unqueried legal features associated with its highest-probability diseases. Balanced quotas retain candidates from both roles. Expected upper-risk reduction ranks candidates within each role and fills unused slots. Ask edges branch according to Equation~\eqref{eq:response-model}. Each simulated response updates the history, observation mask, and both beliefs, then recomputes the protected set and upper risk. When outcomes exceed the branch budget, the most probable outcomes are retained and the remaining mass forms a tail branch.

\paragraph{Search selection.}
Actions are scored by
\begin{equation}
S(s,a)
=
Q(s,a)
+
\beta P(s,a)
\frac{\sqrt{N(s)}}{1+N(s,a)},
\label{eq:puct}
\end{equation}
where $Q(s,a)$ is the estimated action return, $N(s)$ and $N(s,a)$ count visits to the state and action edge, respectively, and $\beta>0$ controls exploration.
\paragraph{Simulation and execution.}
At turn $t$, search uses depth $H_{\mathrm{eff}}=\min(H,T_{\max}-t)$ and up to $B$ simulations. Unvisited Ask actions are explored in descending prior order; subsequent selection uses Equation~\eqref{eq:puct}. Simulations sample response branches, update both beliefs, and back up the return in Equation~\eqref{eq:trajectory-return}. Query cost is charged once, and terminal risk at Commit or the depth limit.

The root executes the highest-value action, breaking value ties by visit count. Search commits when no legal query or no remaining depth is available. With one step remaining, response branches are enumerated directly.

\subsection{Action Execution}
\label{app:execution}

The DDXPlus interface maps each query to an evidence identifier and retrieves the response from the patient record. The observation records feature presence and any categorical values, which enter the belief update in Equation~\eqref{eq:belief-update}. Appendix~\ref{app:mediq-protocol} describes the MediQ language interface.

\subsection{Intake Scoring Prompt}
\label{app:ddx-prompt}
The chat template below supplies the intake, observed finding, and ordered candidate list. Braced fields are populated from the case. Diagnosis names are scored as continuations after the model's assistant generation prefix.

\begin{promptbox}{DDXPlus: System Prompt}
You are a medical diagnostic classifier. Return only the most likely diagnosis name from the supplied list.
\end{promptbox}
\begin{promptbox}{DDXPlus: User Prompt}
Patient: \{initial\_context\}

Known evidence:\newline
- \{question\}: \{answer\}

Candidate diagnoses:\newline
- \{diagnosis\_1\}\newline
...\newline
- \{diagnosis\_49\}

Diagnosis:
\end{promptbox}

%% file: sections/calibration_appendix.tex
\section{Calibration Procedure and Guarantee}
\label{app:conformal}

\subsection{Calibration States}
Calibration proceeds in two stages on disjoint held-out cohorts of 441 cases each: threshold selection, followed by residual calibration with the thresholds fixed. Both cohorts are separate from the test data. Histories sample uniformly from legal, unqueried findings and retrieve responses from the patient record. We record states at turns $t\in\{0,\ldots,10\}$, carrying the last state forward when no legal query remains. Pooling across cases gives $n_t=441$ residuals per turn and backbone.

\subsection{Threshold Selection and Residual Correction}
\paragraph{Differential threshold.}
At each turn, threshold selection uses the safety-belief sets $\{y:b_t^S(y)>\tau\}$ with their highest-probability disease always included. We search 201 equally spaced thresholds from zero to the largest safety probability in the selection sample, together with 0.01. The selected threshold is the largest for which the mean unweighted reference mass omitted plus $\sqrt{\log(1/\delta)/(2n)}$ is at most $\alpha$, with $n=441$ and $\delta=\alpha=0.1$; the fallback threshold is zero. We then freeze this threshold and use the diagnostic leader in the protected set of Equation~\eqref{eq:protected-set}. Residual calibration evaluates that final set and its composite loss on the second subset.

\paragraph{Residual correction.}
For $n_t$ calibration cases at turn $t$, set $k_t=\lceil(n_t+1)(1-\alpha)\rceil$. This gives
\begin{equation}
\widehat q_t=E_{(k_t)}^{(t)},
\label{eq:conformal-quantile}
\end{equation}
where $E_{(k_t)}^{(t)}$ is the $k_t$-th smallest calibration residual. Residuals lie in $[-1,1]$; if $k_t=n_t+1$, we set $\widehat q_t=1$. With $n_t=441$ and $\alpha=0.1$, the correction uses order statistic $k_t=398$. Equation~\eqref{eq:upper-risk} keeps the correction nonnegative and clips the resulting estimate at one, preserving the upper-bound direction and the loss range.

\subsection{Fixed-Turn Guarantee}
Fix a turn $t$ and suppose its $n_t$ calibration residuals and a fresh residual $E_{n_t+1}^{(t)}$ are exchangeable. Their rank symmetry, with ties handled conservatively, gives
\begin{equation*}
\Pr\!\left\{E_{n_t+1}^{(t)}\leq\widehat q_t\right\}
\geq\frac{k_t}{n_t+1}\geq1-\alpha.
\end{equation*}
On this event,
\begin{equation*}
\begin{aligned}
L_t^{\mathrm{test}}
&\leq R_t^{\mathrm{test}}+\widehat q_t\\
&\leq R_t^{\mathrm{test}}+\max(0,\widehat q_t).
\end{aligned}
\end{equation*}
Because $L_t^{\mathrm{test}}\leq1$, clipping the right-hand side at one preserves the inequality and establishes Equation~\eqref{eq:coverage}. The probability is marginal over the calibration sample and the fresh case at the fixed turn.

\paragraph{Policy evaluation.}
Table~\ref{tab:ablation-study} reports Empirical Risk Exceedance for the Ask/Commit policy at its selected stopping turns.

%% file: sections/ddx_appendix.tex
\section{DDXPlus Experimental Details}
\label{app:ddx-setup}
\begingroup
\setlength{\emergencystretch}{1em}
\subsection{Data and Interaction Protocol}
The main evaluation uses 490 cases from the official DDXPlus test split under its natural disease distribution~\cite{tchango2022ddxplus}. The label space contains 49 pathologies; 115 cases have severe primary diagnoses (severity levels 1--2). Each interview begins with age, sex, and one positive finding. Questions reveal unobserved symptoms or antecedents from the patient record. At commitment, the agent returns a leading diagnosis and differential. All methods receive the same intake and legal-question interface. The question budget is ten, with a four-question limit for ATD.

\subsection{Models and Baseline Adaptations}
We evaluate each system with Qwen3-4B, Mistral-7B-Instruct-v0.3, and Llama-3.1-8B-Instruct. The comparison includes an intake-only reference and six active methods. Their DDXPlus implementations are summarized below.

\begin{itemize}
\renewcommand{\labelitemi}{$\bullet$}
\setlength{\itemsep}{3pt}
\setlength{\parsep}{0pt}
\setlength{\topsep}{4pt}
\item \textbf{Direct Commit} returns its diagnosis from the intake, establishing performance before additional evidence is acquired.
\item \textbf{ATD}~\cite{kobalczyk2025active} selects questions that distinguish candidate hypotheses. The adapter maps hypotheses to diagnoses and binary queries to legal clinical findings.
\item \textbf{UoT}~\cite{hu2024uncertainty} uses question partitions estimated from training data and a diagnostic head conditioned on the backbone.
\item \textbf{MISQ-HF}~\cite{chopra2025feedback} combines backbone-conditioned legal question proposals with MCTS and candidate-terminal stopping.
\item \textbf{C-IP}~\cite{chan2025conformal} ranks legal questions by a conformal prediction-set entropy bound and commits when query scores converge.
\item \textbf{ACTMED}~\cite{estevez2025actmed} uses expected multi-disease KL divergence for question selection and a KL-based \mbox{rule for commitment}.
\item \textbf{BED-LLM}~\cite{choudhury2026bed} selects questions by expected information gain and commits when a single hypothesis remains. The DDXPlus adapter supplies recorded clinical findings as patient responses.
\end{itemize}

\subsection{Configuration and Data Separation}
Structured likelihoods are fitted on 100,000 training cases. Two separate 441-case cohorts select differential thresholds and calibrate residuals; a 490-case validation cohort selects hyperparameters. The main test cohort (490 cases) and acquisition-analysis cohort (980 cases) are disjoint samples from the official test split. All these cohorts are mutually disjoint.

Experiments use four NVIDIA RTX A5000 GPUs and the configuration in Table~\ref{tab:ddx-config}, frozen before testing. Main results average five search seeds for CaRDI and use each baseline's recorded run.

\begin{table}[t]
\centering
\small
\begin{tabular*}{\linewidth}{@{\extracolsep{\fill}}lr@{}}
\toprule
Parameter & Value \\
\midrule
Search depth $H$ & 2 \\
Search simulations $B$ & 200 \\
Candidate questions $K_q$ & 8 \\
Proposal diseases per belief & 8 \\
Response branch limit & 12 \\
PUCT coefficient $\beta$ & 1.4 \\
Prior floor $\varepsilon$ & $10^{-4}$ \\
Question cost $c(q)$ & 0.02 \\
Question budget $T_{\max}$ & 10 \\
Intake fusion weight $\eta$ & 0.02 \\
Calibration level $\alpha$ & 0.1 \\
Diagnostic loss weight $\lambda$ & 0.5 \\
Severe omission weight & 2 \\
\bottomrule
\end{tabular*}
\caption{CaRDI configuration for DDXPlus.}
\label{tab:ddx-config}
\end{table}
\endgroup

\subsection{Evaluation Metrics and Output Conventions}
\label{app:ddx-equal-size}

All metrics use the prediction returned at commitment. Let $T_i$ be the number of additional questions for case $i$, $\Gamma_i$ its returned differential, $D_i$ its reference-differential support, and $\mathcal{S}$ the set of severe diseases.

\paragraph{Diagnostic quality and interview length.}
Top-1 and Top-3 measure primary-diagnosis accuracy. DDx-F1 measures reference-differential agreement, and Severity-F1 is the harmonic mean of severe-disease rule-in and rule-out rates. Queries averages $T_i$ across cases.

\paragraph{Clinical loss.}
ECCL averages the realized loss in Equation~\eqref{eq:commit-loss}, using $\lambda=0.5$ and omission weights $\kappa(y)=2$ for severe diseases and $1$ otherwise. Reference weights $g_i$ retain DDXPlus entries above 0.01, assign the primary diagnosis weight 0.01 if absent, and renormalize.

\paragraph{Severe-disease outcomes.}
Severe Error is the rate of incorrect leading diagnoses among the 115 severe-primary cases. Severe Omission is the proportion of $n=490$ cases whose differential omits at least one severe reference alternative:
\begin{equation*}
\frac{1}{n}\sum_{i=1}^{n}
\mathbf{1}\!\left\{(D_i\cap\mathcal{S})\setminus\Gamma_i\neq\varnothing\right\}.
\end{equation*}
\paragraph{Risk exceedance.}
Empirical Risk Exceedance is the proportion of cases whose realized loss exceeds the reported risk at commitment:
\begin{equation*}
\frac{1}{n}\sum_{i=1}^{n}
\mathbf{1}\!\left\{L_{i,T_i}>R^{\mathrm{reported}}_{i,T_i}\right\}.
\end{equation*}
Reported risk is $U$ for CaRDI and plug-in $R$ for the variant w/o Conformal Calibration.

\paragraph{Native differential.}
CaRDI uses the threshold-selected differential in Equation~\eqref{eq:protected-set} for planning and evaluation. Section~\ref{sec:ddx-main-results} reports mean sizes.

\paragraph{Common output capacity.}
The external-system comparison also evaluates the first three entries of each stored terminal diagnostic ranking. The leading prediction, acquired evidence, and stopping time are preserved.

The first-question and four-question controls evaluate the diagnostic leader and four highest-ranked safety alternatives. Question selection uses the native threshold-based differential; the five-label evaluation measures the effect of acquired evidence at a common output size.

%% file: sections/analysis_appendix.tex
\section{Additional Results and Analyses}
\label{app:analysis}

\subsection{Shared Protocol}
The controlled analyses isolate question selection, calibration, severity weighting, and differential retention. In the question-selection comparisons, risk and entropy planners share training data, both structured likelihoods, cached Llama-3.1-8B intake scores, response models, candidate-generation rules, and search budgets. Their planning objectives differ. These comparisons retain five diagnoses: the diagnostic leader and four highest-ranked safety alternatives.

\paragraph{Statistical comparisons.}
Paired comparisons average search seeds within each patient and use 5,000 patient-bootstrap resamples for 95\% confidence intervals, except the output-capacity analysis, which uses 2,000. The following experiments specify their cohorts and seeds.

\subsection{Question Selection under Fixed Budgets}
\paragraph{First question.}
Both planners use depth two, 200 simulations, and paired seeds 0--2 on the 980-case acquisition cohort, which is disjoint from the main test cohort. Evaluation follows the first observed response. Risk-guided planning improves accuracy by 2.99 percentage points (95\% CI $1.53$ to $4.46$) and changes ECCL by $-0.0187$ ($-0.0279$ to $-0.0098$). The normalized posterior-entropy difference is $-0.0011$ ($-0.0082$ to $0.0061$). First questions differ in 60.7\% of cases.

\paragraph{Intake uncertainty.}
Validation-derived entropy terciles (0.4091 and 0.7183, from 490 intake states) divide the acquisition cohort into 352 low-, 290 middle-, and 338 high-entropy patients. Accuracy gains are 1.52, 0.34, and 6.80 points; ECCL differences are $-0.0095$, $+0.0004$, and $-0.0445$, respectively. The high-entropy group's accuracy gain exceeds that of the remaining patients by 5.82 points (95\% CI $2.36$ to $9.33$).

\paragraph{Four-question extension.}
The four-question comparison uses the 490 main-test patients and paired seeds 0--2. Both planners use depth four, 800 simulations, and eight candidate questions, selected on the separate validation cohort. Evaluation retains five diagnoses. Risk-guided planning reduces the severe-disease contribution to weighted omission by 0.0069 (95\% CI for risk minus entropy: $-0.0137$ to $-0.0005$). This contribution is the portion of Equation~\eqref{eq:true-omission} attributable to severe omitted diseases.

\subsection{Calibration Replay}
Replay isolates how residual corrections affect Ask/Commit decisions. Sorting SHA-256 hashes of case identifiers selects 240 acquisition-cohort patients. Their seed-zero trajectories contain 205 pre-final-question states and 221 commitment states before the question limit. We hold beliefs, histories, thresholds, seed, depth two, and 200 simulations fixed, then set the composite residual correction to zero. Replaying the calibrated setting reproduces all 426 recorded actions.

Exact one-step enumeration uses the common legal proposal pool; exact two-step search uses the original eight-question shortlist. At one step, a common successor correction preserves question rankings when clipping is inactive. At two steps, branches can commit at different turns and incur different corrections. Setting the correction to zero changes 100/205 pre-final MCTS actions and 17/205 exact two-step actions, illustrating the interaction between calibration and stopping.

\subsection{Severity Weighting}
This comparison tests the effect of assigning higher omission costs to severe diseases. CaRDI and a uniform-cost variant, $\kappa(y)=1$, use the same 490 main-test patients, paired seeds 0--2, native protected differentials, and adaptive stopping. Each uses its corresponding residual calibration. Evaluation applies the common twofold severe-disease costs in Equation~\eqref{eq:true-omission}. The severe contribution uses that equation's total consequence-weighted reference mass as its denominator.

Severity weighting reduces this contribution from 0.04391 to 0.04018, an 8.5\% relative reduction. The paired difference is $-0.00373$ (95\% CI $-0.00756$ to $-0.00004$). Both variants achieve 96.39\% Top-1 accuracy, with 3.585 questions for severity weighting and 3.581 for uniform costs.

Native ECCL is 0.05819 versus 0.06049, with paired difference $-0.00230$ ($-0.00745$ to $0.00287$). Severe Omission is 29.86\% versus 30.68\%, with difference $-0.82$ percentage points ($-2.52$ to $0.88$). Mean differential sizes are 15.17 and 14.98, respectively.

\subsection{Retention at Fixed Output Capacity}
\label{app:output-capacity}
\begin{figure*}[t]
\centering
\includegraphics[width=\textwidth]{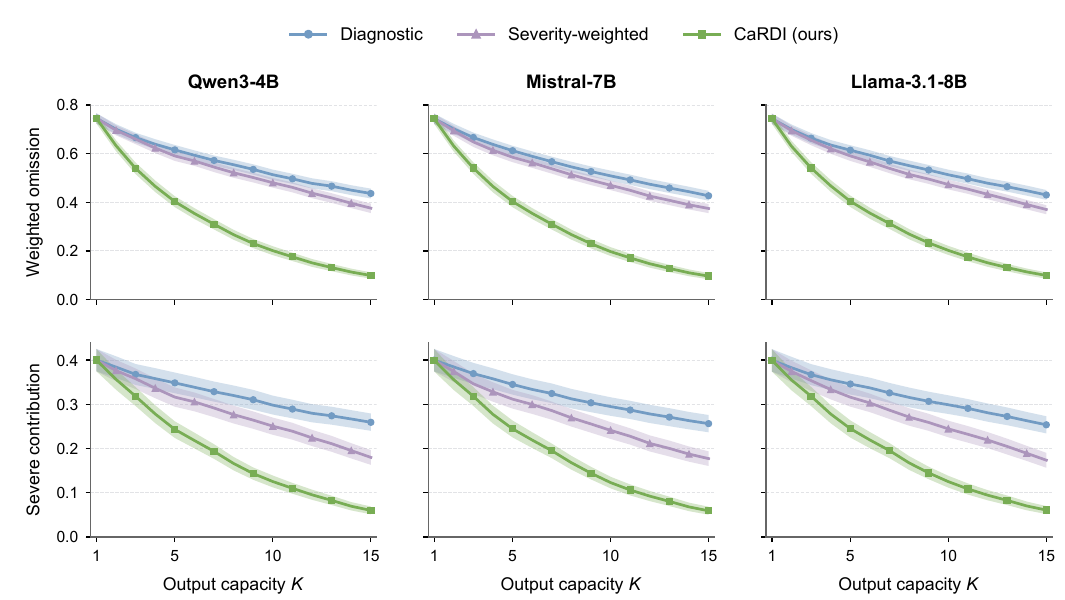}
\caption{Differential retention across backbones and output capacities. Top: weighted omission; bottom: the contribution from severe diseases. Rankings use identical evidence and primary predictions. Shaded bands show 95\% patient-bootstrap confidence intervals.}
\label{fig:output-capacity-full}
\end{figure*}

We use the original 490 main-test trajectories for each backbone and all five search seeds. At each terminal state, all three output rules retain the original primary prediction. The remaining $K-1$ diagnoses are selected by diagnostic belief $b^D(y)$, severity-weighted diagnostic scores $\kappa(y)b^D(y)$, or safety belief $b^S(y)$ for CaRDI, with $K=1,\ldots,15$. We use the same severity weights as the clinical loss: two for severe diseases and one otherwise. Evidence acquisition and stopping remain fixed. Curves average seeds within patients, then across patients; confidence bands in Figure~\ref{fig:output-capacity} use 2,000 patient-bootstrap resamples.

At five diagnoses per output, safety-belief ranking reduces weighted omission by 31.1--31.8\% relative to severity-weighted diagnostic ranking (Table~\ref{tab:retention-five}). Figure~\ref{fig:output-capacity-full} gives the full capacity curves.

\begin{table}[!htb]
\centering
\small
\setlength{\tabcolsep}{3pt}
\begin{tabular*}{\linewidth}{@{\extracolsep{\fill}}lrrr@{}}
\toprule
Ranking & Qwen & Mistral & Llama \\
\midrule
\multicolumn{4}{@{}l}{\textit{Weighted omission}} \\
Diagnostic & 0.615 & 0.612 & 0.614 \\
Severity-weighted & 0.591 & 0.585 & 0.591 \\
Safety (CaRDI) & 0.403 & 0.403 & 0.404 \\
\midrule
\multicolumn{4}{@{}l}{\textit{Severe-disease contribution}} \\
Diagnostic & 0.348 & 0.345 & 0.346 \\
Safety (CaRDI) & 0.243 & 0.245 & 0.245 \\
\midrule
\multicolumn{4}{@{}l}{\textit{Omission reduction vs. severity-weighted}} \\
Safety (CaRDI) & 31.8\% & 31.1\% & 31.6\% \\
\bottomrule
\end{tabular*}
\caption{Differential retention with five diagnoses per output. Rankings share terminal evidence and primary predictions; results average five seeds per patient over 490 DDXPlus cases.}
\label{tab:retention-five}
\end{table}

Relative to diagnostic ranking, paired 95\% confidence intervals for the reduction in severe contribution are $0.0912$ to $0.1190$ for Qwen, $0.0875$ to $0.1136$ for Mistral, and $0.0875$ to $0.1141$ for Llama. The contribution uses the reference-mass normalization in Equation~\eqref{eq:true-omission}.

\subsection{Paired Comparisons with External Systems}
\label{app:main-paired}
The main-system comparisons use the same 490 patients for both methods. For each patient, CaRDI averages its five recorded seeds; the baseline uses its recorded main-table trajectory. Against BED-LLM, Top-1 gains across Qwen/Mistral/Llama are 8.78/9.35/9.02 percentage points, with respective 95\% confidence intervals $5.96$ to $11.59$, $6.57$ to $12.20$, and $6.29$ to $11.84$. Query differences are $-0.40/-0.50/-0.47$, with intervals $-0.71$ to $-0.10$, $-0.81$ to $-0.19$, and $-0.77$ to $-0.16$.

\paragraph{Equal output capacity.}
At three diagnoses per output, CaRDI lowers ECCL relative to ACTMED by 12.6--13.4\% across backbones. The improvement comes from higher primary-diagnosis accuracy. Paired patient-bootstrap intervals for ACTMED-minus-CaRDI ECCL are $0.0353$ to $0.0665$, $0.0354$ to $0.0680$, and $0.0379$ to $0.0703$ for Qwen, Mistral, and Llama, respectively.

\subsection{Illustrative First-Question Comparison}
Case \emph{test:6218} illustrates how the two objectives value the same intake evidence. The comparison uses the recorded seed-zero trajectories from the first-question study.

\begin{casebox}{DDXPlus: First-Question Comparison}
\textbf{Initial evidence}\par
``Do you have a cough?'' --- ``Yes.''

\textbf{Reference diagnosis}\par
Viral pharyngitis.

\textbf{Risk-guided question}\par
``Have you had significantly increased sweating?''\par
Response: ``No.''\par
Leading diagnosis after the response: Viral pharyngitis.\par
Normalized entropy: $0.718\rightarrow0.723$.

\textbf{Entropy-guided question}\par
``Do you have pain somewhere, related to your reason for consulting?''\par
Response: ``Yes.''\par
Leading diagnosis after the response: URTI.\par
Normalized entropy: $0.718\rightarrow0.624$.
\end{casebox}

The risk-guided question yields the correct leading diagnosis, while the entropy-guided question produces a larger entropy reduction.

%% file: sections/mediq_protocol.tex
\section{MediQ Experimental Details}
\label{app:mediq-protocol}
\subsection{Language Interface}
MediQ pairs an Expert with a Patient to acquire evidence for a four-option medical question. Each episode starts with the intake, question, and answer options. A cached question bank links natural-language queries to individual clinical facts. The Patient deterministically returns the fact associated with the selected query. The Expert then scores the options using the intake and accumulated responses. Queried facts leave the pool.

\subsection{Simulated Belief Transitions}
Search represents a belief by its maximum option probability, normalized entropy, and answer-set size. Empirical transitions are estimated from training and calibration histories, grouped by the clinical question type derived from query wording. Transitions condition on the current turn and belief summary, with pooled question-type distributions providing backoff.

Each sampled successor creates a response branch. Its maximum option probability and turn-specific residual correction determine the Commit value. Executing the chosen query supplies the actual Patient response, after which the Expert scores all four options afresh.

\subsection{Configuration}
The Expert uses Llama-3.1-8B-Instruct with sequence-probability scoring. Commit loss is answer error ($\lambda=1$), estimated by one minus the highest option probability. Search uses depth three, 512 simulations, and a transition-sampling budget of 800. Query cost is zero, returns are undiscounted, and the interview allows zero to ten queries. The evaluation covers folds zero through two with seed zero.

\subsection{Calibration and Metrics}
\paragraph{Residual calibration.}
Random-query histories provide up to 200 distinct calibration cases per turn, with $\alpha=0.1$. The planner evaluates commitment using calibrated answer-error risk.

\paragraph{Answer-set coverage.}
For scored option probabilities $p_t(y)$ after $t$ questions, the answer set is
\begin{equation*}
\mathcal{S}_t=\{y:1-p_t(y)\leq\widehat{\tau}_t\}.
\end{equation*}
The threshold $\widehat{\tau}_t$ is calibrated from reference-option scores $1-p_t(y^*)$ on held-out random-query histories at the same turn, with $\alpha=0.1$. It is fitted to the probability scores used by the evaluated method. Coverage measures inclusion of $y^*$ in $\mathcal{S}_T$ at the final turn $T$.

\paragraph{Accuracy and interaction.}
Accuracy measures final top-option correctness; Queries counts additional questions. Accuracy--turn AUC@10 integrates accuracy from turns zero through ten by the trapezoidal rule and divides by ten, carrying stopped predictions forward. Full-Info Gap is All Information accuracy minus the method's accuracy, in percentage points.

\subsection{Answer Scoring Prompt}
\label{app:mediq-prompt}
The Expert uses the following system and user messages. Braced fields are populated from the case and observed conversation. An assistant prefix \verb|{"answer": "| precedes the candidate answer label for probability scoring.

\begin{promptbox}[before skip=3pt,after skip=3pt,top=4pt,bottom=4pt]{MediQ: System Prompt}
You are a medical doctor answering real-world medical entrance exam questions. Based on your understanding of basic and clinical science, medical knowledge, and mechanisms underlying health, disease, patient care, and modes of therapy, answer the following multiple choice question. Select one correct answer from A to D. Base your answer on current standard medical practice. Reason through the clinical evidence internally, then respond with only the answer field in this format:

\{"answer": "A/B/C/D"\}
\end{promptbox}
\begin{promptbox}[before skip=3pt,after skip=3pt,top=4pt,bottom=4pt,before upper={\raggedright\setlength{\parskip}{2pt}}]{MediQ: User Prompt}
Answer the multiple choice based on the context.

Initial Info: \{init\_info\}

Conversation Log between doctor and patient:\newline
\{history\_lst\}

Question: \{question\}\newline
Options:\newline
    A - \{option\_a\}\newline
    B - \{option\_b\}\newline
    C - \{option\_c\}\newline
    D - \{option\_d\}

Please select the most appropriate answer (A/B/C/D).

Medical conditions are complex. First identify the relevant specialty, then compare the findings that support or contradict each option, including important missing evidence. Think step by step about the patient information and the options, but return only the requested answer JSON.
\end{promptbox}

\newpage
\subsection{Illustrative Interaction}
\label{app:mediq-case}
In development case \emph{all/split2/train/209}, held out in fold two, an imaging response changes the leading answer and supports commitment after one question. Both methods use the same intake, question bank, backbone, and recorded responses.

\begin{casebox}{MediQ: From Evidence to Commitment}
\textbf{Initial presentation}\par
A 58-year-old man presents with four hours of severe abdominal pain.

\textbf{Question and options}\par
Which of the following is the best next step in the management of this patient?\par
A: Cholecystectomy\newline
B: Colonoscopy\newline
C: Endoscopic retrograde cholangiopancreatography (ERCP)\newline
D: Esophagogastroduodenoscopy (EGD)

\textbf{Initial answer belief}\par
EGD leads with probability 0.379; ERCP has probability 0.198.

\textbf{Selected question}\par
``What additional imaging information is available?''

\textbf{Observed response}\par
``CT of the abdomen shows an enlarged gallbladder and peripancreatic fluid with dilation of the pancreatic and common bile ducts''

\textbf{Decision}\par
The probability of ERCP rises to 0.944. CaRDI commits to ERCP after one question, matching the reference answer. Entropy-Greedy asks four questions and selects EGD.
\end{casebox}